\documentclass[english]{article}
\usepackage[T1]{fontenc}
\usepackage{amsmath}
\usepackage{amsthm}
\usepackage{amssymb}
\usepackage{graphicx}

\makeatletter
\theoremstyle{plain}
\newtheorem{thm}{\protect\theoremname}

\usepackage{amsfonts}
\usepackage{ml2k}
\usepackage{chicago}

\makeatother

\usepackage{babel}
\providecommand{\theoremname}{Theorem}

\begin{document}
\twocolumn[

\icmltitle{A PAC-Bayesian Analysis of Distance-Based Classifiers:\\ Why Nearest-Neighbour works!}

\icmlauthor{Thore Graepel}{thoregraepel@gmail.com}

\icmlauthor{Ralf Herbrich}{ralf@herbrich.me}

\icmladdress{} \vskip 0.3in ] 
\begin{abstract}
We present PAC-Bayesian bounds for the generalisation error of the
$K$-nearest-neighbour classifier ($K$-NN). This is achieved by casting
the $K$-NN classifier into a kernel space framework in the limit
of vanishing kernel bandwidth. We establish a relation between prior
measures over the coefficients in the kernel expansion and the induced
measure on the weight vectors in kernel space. Defining a sparse prior
over the coefficients allows the application of a PAC-Bayesian folk
theorem that leads to a generalisation bound that is a function of
the number of redundant training examples: those that can be left
out without changing the solution. The presented bound requires to
quantify a prior belief in the sparseness of the solution and is evaluated
after learning when the actual redundancy level is known. Even for
small sample size ($m\approx100$) the bound gives non-trivial results
when both the expected sparseness and the actual redundancy are high.
\end{abstract}

\section{Introduction}

The $K$-\emph{nearest}-\emph{neighbour} ($K$-NN) \cite{FixHodges:51,FixHodges:52}
classifier is an elegantly simple and surprisingly effective learning
machine. It takes as input a set of training objects and their labels,
and returns for a given test object represented in terms of its \emph{pairwise}
\emph{distances} to the training objects a label that is determined
by a majority vote over the labels of the $K$ nearest neighbours
in the training sample. $K$-NN is not only conceptually simple but
also very versatile because it does not require a vectorial representation
but only the pairwise distances between test and training objects.
It is thus applicable to all kinds of \emph{structural} \emph{data}
like strings or graphs as long as a meaningful (in the sense of the
classification task) distance measure can be defined. $K$-NN also
has some remarkable asymptotic properties. It is \emph{universally}
\emph{consistent} in the sense that it converges to the Bayes decision
if $K\rightarrow\infty$ and $K/m\rightarrow0$ as the training sample
size $m\rightarrow\infty$. Also under certain regularity conditions
the risk of $1$-NN for $m\rightarrow\infty$ is bounded from above
by twice the Bayes error, $R_{\infty}\left(\mathrm{NN}_{1}\right)\leq2R\left(h_{\mathrm{Bayes}}\right)$,
while for $K$-NN it can be shown that $R_{\infty}\left(\mathrm{NN}_{K}\right)\leq R\left(h_{\mathrm{Bayes}}\right)(1+\sqrt{2/K})$.
With regard to the \emph{computational} \emph{effort} a simple analysis
yields $\mathcal{O}\left(mKd\right)$ where $d$ represents the cost
of one distance evaluation. More refined analysis reveals that for
fixed $K$ and $d$ the worst case time is $\mathcal{O}\left(m^{1/d}\right)$
and the expected time is $\mathcal{O}\left(\log m\right)$. These
results and more regarding $K$-NN can be found in \cite{Devroye:96}.

In this paper we will be concerned with bounds on the risk of the
$K$-NN classifier for \emph{small} \emph{sample} \emph{size}. Why
should such an analysis be of interest? To answer this question consider
the infamous \emph{no}-\emph{free}-\emph{lunch} \emph{theorem} by
Wolpert \cite{Wolpert:95}. This theorem essentially states that averaged
over a uniform distribution over all learning problems no classifier
is better than any other. This theorem may at first glance leave no
hope for the successful development of reliable learning algorithms.
More careful analysis, however, reveals that only the objective of
developing a universally best learning machine is led ad absurdum.
What the theorem \emph{does} tell us is that given a sample and a
learning algorithm we should require the learning algorithm to output
not only a classifier but also a \emph{performance} \emph{guarantee}:
we require the learning algorithm to be \emph{self}-\emph{bounding}
\cite{Freund:98}. This performance guarantee is best given in terms
of an \emph{a-posteriori} bound on the risk of the classifier. Standard
PAC/VC theorems provide \emph{a-priori} results in the sense that
the bound value is entirely determined by the level of confidence
$1-\delta$, the number $m$ of training examples, the empirical risk
$R_{\mathrm{emp}}$, and the complexity of the hypothesis class $\mathcal{H}$
--- usually expressed in terms of its VC-dimension $d_{\mathrm{VC}}$.
These bounds can thus be evaluated \emph{before} learning if $R_{\mathrm{emp}}=0$
is enforced, or after learning when $R_{\mathrm{emp}}$ is known.
In contrast, an a-posteriori bound may only be evaluated \emph{after}
learning, because it takes into account the match between the hypothesis
class $\mathcal{H}$ and the training data $Z$, e.g.~in terms of
the \emph{margin} observed on the training sample.

The idea of a-posteriori bounds was developed in statistical learning
theory and the first conceptual framework for such bounds was \emph{structural
risk minimisation} \cite{Vapnik:98}\emph{.} The idea was further
developed to include \emph{data}-\emph{dependent} structural risk
minimisation \cite{Taylor:96} that is capable of exploiting \emph{luckiness}
w.r.t.~the match of input data and learning machine. The latest results
are now known as the \emph{PAC}-\emph{Bayesian} \emph{framework} based
on work by David McAllester \cite{McAllester:98}. Note that the PAC-Bayesian
framework also provided the basis for the discovery of a very tight
margin bound for linear classifiers in kernel spaces \cite{Herbrich:99b}.

In Section \ref{sec:pac_bayes} we introduce basic concepts and notation
and present the PAC-Bayesian results on which our analysis is based.
In Section \ref{sec:knn_review} we briefly review the definition
of the $K$-nearest-neighbour classifier. In Section \ref{sec:1nn}
the $1$-NN algorithm is formulated as the limiting case of a linear
classifier in a kernel space. This leads to an intuitive explanation
of its generalisation ability. The resulting hypothesis space is then
used in Section \ref{sec:bound} by defining a sparse prior that leads
to a PAC-Bayesian bound for $1$-NN. This result will is generalised
to $K$-NN in Section \ref{sec:knn}. Finally, we conclude and point
to ideas for future work by relating $K$-NN to Support Vector Machines
(SVM) \cite{Vapnik:98}.

Throughout the paper we denote probability measures by $\boldsymbol{\mathsf{P}}_{\mathsf{H}}$
and the related expectation by $\boldsymbol{\mathsf{E}}_{\mathsf{H}}$.
The subscript refers to the random variable. 

\section{Learning in the PAC-Bayesian framework\label{sec:pac_bayes}}

We consider the learning of binary classifiers. We define learning
as the process of selecting one hypothesis $h$ from a given hypothesis
space $\mathcal{H}$ of hypotheses $h:\mathcal{O}\rightarrow\mathcal{Y}$
that map objects $\boldsymbol{\mathsf{o}}\in\mathcal{O}$ to labels
$y\in\mathcal{Y}=\left\{ -1,+1\right\} $. The selection is based
on a training sample $Z$ comprised of a set $O=\left\{ \boldsymbol{\mathsf{o}}_{1},\ldots,\boldsymbol{\mathsf{o}}_{m}\right\} \in\mathcal{O}^{m}$
of objects and their corresponding labels. We will assume the training
sample $Z$ to be drawn iid from a probability measure $\boldsymbol{\mathsf{P}}_{\mathsf{Z}}\equiv\boldsymbol{\mathsf{P}}_{\mathsf{OY}}=\boldsymbol{\mathsf{P}}_{\mathsf{Y}|\mathsf{O}}\boldsymbol{\mathsf{P}}_{\mathsf{O}}$.
Based on these definitions let us define the risk $R\left(h\right)$
of a hypothesis $h$ by 
\[
R\left(h\right)=\boldsymbol{\mathsf{P}}_{\mathsf{Z}}\left[h\left(\mathsf{O}\right)\neq\mathsf{Y}\right]\,.
\]
A reasonable criterion for learning is to try to find the the hypothesis
$h^{*}=\mathrm{argmin}_{h}R\left(h\right)$ that minimises the risk.
The difficulty in this learning task lies in the fact that the probability
measure $\boldsymbol{\mathsf{P}}_{\mathsf{Z}}$ is unknown. Let us
define the empirical risk $R_{\mathrm{emp}}\left(h,Z\right)$ of an
hypothesis $h\in\mathcal{H}$ on a training sample $Z$ by
\begin{equation}
R_{\mathrm{emp}}\left(h,Z\right)=\frac{1}{m}\left|\left\{ \left(\boldsymbol{\mathsf{o}},y\right)\in Z:h\left(\boldsymbol{\mathsf{o}}\right)\neq y\right\} \right|\,.\label{eq:r_emp}
\end{equation}
 The principle of \emph{empirical risk minimisation \cite{Vapnik:98}}
advocates minimising the empirical risk $R_{\mathrm{emp}}\left(h,Z\right)$
instead of the true risk $R\left(h\right)$. 

An a-posteriori bound aims at bounding the risk $R\left(h\right)$
of an hypothesis $h$ based on the knowledge of $\mathcal{H}$ as
well as $Z$. We now present two theorems by D.~McAllester \cite{McAllester:98}
that require the definition of a prior measure $\boldsymbol{\mathsf{P}}_{\mathsf{H}}$
on $\mathcal{H}$ and reward the selection of a hypothesis of high
prior weight with a low bound on the generalisation error. Note that
these theorems do not depend on the correctness of $\boldsymbol{\mathsf{P}}_{\mathsf{H}}$.
If the belief expressed in $\boldsymbol{\mathsf{P}}_{\mathsf{H}}$
turns out to be wrong, the bounds just become trivial.
\begin{thm}
\label{thm:dma1}For any probability measure $\boldsymbol{\mathsf{P}}_{\mathsf{H}}$
over an hypothesis space $\mathcal{H}$ containing a target hypothesis
$h^{*}\in\mathcal{H}$, and any probability measure $\boldsymbol{\mathsf{P}}_{\mathsf{Z}}$
on labelled objects, we have, for any $\delta>0,$ that with probability
at least $1-\delta$ over the selection of a sample $Z$ of $m$ examples,
the following holds for all hypotheses $h\in\mathcal{H}$ agreeing
with $h^{*}$on that sample:
\[
R\left(h\right)\leq\frac{\log\frac{1}{\boldsymbol{\mathsf{P}}_{\mathsf{H}}\left(h\right)}+\log\frac{1}{\delta}}{m}\,.
\]
\end{thm}
To see that this is true note that the probability that a hypothesis
$h$ with risk $R\left(h\right)$ is consistent with a sample of $m$
examples is bounded from above by $\left(1-R\left(h\right)\right)^{m}\leq\exp\left(-mR\left(h\right)\right)$.
If $R\left(h\right)$ is greater than the above bound the probability
that $h$ is consistent with the sample is bounded from above by $\boldsymbol{\mathsf{P}}_{\mathsf{H}}\left(h\right)\delta.$
Applying the union bound the probability that some hypothesis $h$
that violates the bound is consistent with the sample is bounded by
$\sum_{h\in\mathcal{H}}\boldsymbol{\mathsf{P}}_{\mathsf{H}}\left(h\right)\delta=\delta$.

Essentially replacing the binomial tail bound used in the above argument
by the Chernoff bound for bounded random variables leads to an agnostic
version of the above Theorem \ref{thm:dma1}.
\begin{thm}
\label{thm:dma2}For any probability measure $\boldsymbol{\mathsf{P}}_{\mathsf{H}}$
over an hypothesis space $\mathcal{H}$, and any probability measure
$\boldsymbol{\mathsf{P}}_{\mathsf{Z}}$ on labelled objects, we have,
for any $\delta>0,$ that with probability at least $1-\delta$ over
the selection of a sample $Z$ of $m$ examples, all hypotheses $h\in\mathcal{H}$
satisfy
\[
R\left(h\right)\leq R_{\mathrm{emp}}\left(h,Z\right)+\sqrt{\frac{\log\frac{1}{\boldsymbol{\mathsf{P}}_{\mathsf{H}}\left(h\right)}+\log\frac{1}{\delta}}{2m}}\,.
\]
\end{thm}
In both theorems the complexity term as found, e.g.~in VC bounds,
is replaced by the negative $\log$-prior of the hypothesis at hand.
Thus if the prior belief in that particular hypothesis was high the
effective complexity is low and the bound gives small values. Before
we can apply these bounds to $K$-NN we need to cast this classifier
into the appropriate framework.

\section{The $K$-nearest-neighbour classifier\label{sec:knn_review}}

The $K$-nearest-neighbour classifier requires that the set $\mathcal{O}$
of objects $\boldsymbol{\mathsf{o}}\in\mathcal{O}$ be equipped with
a distance measure. Although not strictly necessary for the application
of $K$-NN we will assume that we have a \emph{metric} $d:\mathcal{O}\times\mathcal{O}\mapsto\mathbb{R}^{+}$
between objects. Then the $K$-nearest-neighbour classifier is a mapping
$\mathrm{NN}_{K}:\left(O\times\mathcal{Y}\right)^{m}\times O\mapsto\mathcal{Y}$
defined as follows,
\begin{eqnarray}
\mathrm{NN}_{K}\left(Z,\boldsymbol{\mathsf{o}}\right) & = & \textrm{sign}\left(\sum_{\stackrel{\left(\boldsymbol{\mathsf{o}}^{\prime},y^{\prime}\right)\in Z:}{\boldsymbol{\mathsf{o}}^{\prime}\in N_{K}(\boldsymbol{\mathsf{o}})}}y^{\prime}\right)\nonumber \\
 & = & \textrm{sign}\left(\sum_{i:\boldsymbol{\mathsf{o}}_{i}\in N_{K}(\boldsymbol{\mathsf{o}})}y_{i}\right),\label{eq:knn}
\end{eqnarray}
 where the $K$-neighbourhood $N_{K}\left(\boldsymbol{\mathsf{o}}\right)$
is defined for $\boldsymbol{\mathsf{o}}^{\prime},\boldsymbol{\mathsf{o}}^{\prime\prime}\in O$
as 
\[
N_{K}\left(\boldsymbol{\mathsf{o}}\right)=\left\{ \boldsymbol{\mathsf{o}}^{\prime}:\left|\left\{ \boldsymbol{\mathsf{o}}^{\prime\prime}:d\left(\boldsymbol{\mathsf{o}},\boldsymbol{\mathsf{o}}^{\prime\prime}\right)<d\left(\boldsymbol{\mathsf{o}},\boldsymbol{\mathsf{o}}^{\prime}\right)\right\} \right|<K\right\} .
\]
Note that this definition may lead to $K$-neighbourhoods of cardinality
$\left|N_{K}\left(\boldsymbol{\mathsf{o}}\right)\right|>K$ in the
case of a distance tie $d\left(\boldsymbol{\mathsf{o}},\boldsymbol{\mathsf{o}}^{\prime}\right)=d(\boldsymbol{\mathsf{o}},\boldsymbol{\mathsf{o}}^{\prime\prime})$
for some $\boldsymbol{\mathsf{o}}^{\prime},\boldsymbol{\mathsf{o}}^{\prime\prime}\in O$.
Let us explicitly break this tie and enforce $\left|N_{K}\left(\boldsymbol{\mathsf{o}}\right)\right|=K$
by discarding those objects in the tie with a higher index. 

Also, for $K$ even there may result a voting-tie in the decision
leading to $\mathrm{NN}_{K}(Z,\boldsymbol{\mathsf{o}})=0$. Of course,
a tie of the latter type may serve as an indicator of an uncertain
prediction. 

The above formulation of $K$-NN reflects the basic algorithm. Extensions
have been suggested (see \cite{Devroye:96}) that allow for different
weighting factors depending on the ranks of the neighbours. For conceptual
clarity, such extensions are not considered here.

\section{The 1-NN classifier as the limit of a kernel classifier\label{sec:1nn}}

Let us first consider the case of $K=1$. Then the definition of the
neighbourhood is reduced to 
\[
N_{1}\left(\boldsymbol{\mathsf{o}}\right)=\left\{ \boldsymbol{\mathsf{o}}^{\prime}\in O:\boldsymbol{\mathsf{o}}^{\prime}=\mathrm{argmin}_{\boldsymbol{\mathsf{o}}^{\prime\prime}\in O}d\left(\boldsymbol{\mathsf{o}},\boldsymbol{\mathsf{o}}^{\prime\prime}\right)\right\} .
\]
 In order to be able to view the NN-classifier as a linear classifier
in a kernel space let us introduce a $\mathrm{softmin}$-function
so as to replace the $\mathrm{argmin}$-function. Since the kernel
used should conform to the Mercer conditions \cite{Mercer:09} in
order to ensure the desirable properties of a kernel space, we leave
the soft-min function unnormalised. This does not change the output
of the classifier under the $\mathrm{sign}$-function and leads to

\begin{eqnarray}
\mathrm{NN}_{1}\left(Z,\boldsymbol{\mathsf{o}}\right) & = & \textrm{sign}\left(\sum_{i=1}^{m}\lim_{\sigma\rightarrow0}k_{\sigma}\left(\boldsymbol{\mathsf{o}},\boldsymbol{\mathsf{o}}_{i}\right)y_{i}\right).\label{eq:1nn}
\end{eqnarray}
We can use any positive definite kernel $k_{\sigma}:\mathcal{O}\times\mathcal{O}\mapsto\mathbb{R}^{+}$
with $k_{\sigma}\left(\boldsymbol{\mathsf{o}},\boldsymbol{\mathsf{o}}^{\prime}\right)=k_{\sigma}\left(d^{2}\left(\boldsymbol{\mathsf{o}},\boldsymbol{\mathsf{o}}^{\prime}\right)\right)$
(satisfying the Mercer conditions) and for which for any countable
set $I\subset\mathbb{R}^{+}$ of positive real numbers we have
\[
\lim_{\sigma\rightarrow0}\frac{k_{\sigma}\left(d\right)}{\sum\limits _{d^{\prime}\in I}k_{\sigma}\left(d^{\prime}\right)}=\left\{ \begin{array}{ll}
1 & \qquad\textrm{ if }d=\min\left(I\right)\\
0 & \qquad\textrm{ otherwise}
\end{array}\right.\,.
\]
 Such a kernel is, e.g.~given by the RBF kernel
\begin{equation}
k_{\sigma}\left(\boldsymbol{\mathsf{o}},\boldsymbol{\mathsf{o}}^{\prime}\right)=\exp\left(-\frac{d^{2}\left(\boldsymbol{\mathsf{o}},\boldsymbol{\mathsf{o}}^{\prime}\right)}{\sigma^{2}}\right)\,,\label{eq:rbf}
\end{equation}
 which we will use in the following. 

\begin{figure}
\begin{centering}
\includegraphics[width=0.9\columnwidth]{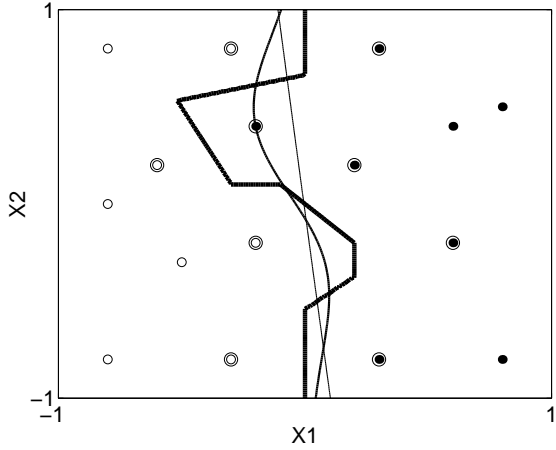}
\par\end{centering}
\caption{\label{fig:1nn}Illustration of the convergence of the classifier
based on class-conditional Parzen window density estimation to the
$1$-NN classifier in $\mathcal{O}=\left[-1,+1\right]^{2}\subset\mathbb{R}^{2}$
using $d^{2}\left(\mathbf{x},\mathbf{x}^{\prime}\right)=\left\Vert \mathbf{x}-\mathbf{x}^{\prime}\right\Vert ^{2}$.
For $\sigma=5$ the decision surface (thin line) is almost linear,
for $\sigma=0.4$ the curved line (medium line) results, and for very
small $\sigma=0.02$ the piecewise linear decision surface (thick
line) of $1$-NN is approaches. For $1$-NN only the circled points
contribute to the decision surface, comparable to support vectors
in the SVM. }
\end{figure}
There exists an interesting relation to the Bayes optimal classifier
$h_{\mathrm{Bayes}}$ that can be approximated using the Parzen window
\cite{Parzen:62} kernel density estimate with kernel $k_{\sigma}$.
If the objects $\boldsymbol{\mathsf{o}}\in\mathcal{O}$ are represented
as vectors $\mathbf{x}_{\boldsymbol{\mathsf{o}}}\in\mathcal{X}\subseteq\mathbb{R}^{n}$,
i.e.~$d\left(\boldsymbol{\mathsf{o}},\boldsymbol{\mathsf{o}}^{\prime}\right)=\left\Vert \mathbf{x}_{\boldsymbol{\mathsf{o}}}-\mathbf{x}_{\boldsymbol{\mathsf{o}}^{\prime}}\right\Vert $
then the class conditional density $\boldsymbol{\mathsf{f}}_{\mathsf{X}|\mathsf{Y}=\pm1}$
can be estimated using a Parzen window density estimator 
\[
\boldsymbol{\widehat{\mathsf{f}}}_{\mathsf{X}|\mathsf{Y}=\pm1}\left(\mathbf{x}\right)=\frac{1}{m^{\pm}}\sum_{i=1}^{m^{\pm}}k_{\sigma}\left(\mathbf{x},\mathbf{x}_{\boldsymbol{\mathsf{o}}_{i}}\right)\,,
\]
where the kernel is assumed to be normalised to one, i.e.~$\int_{\mathcal{X}}k_{\sigma}\left(\mathbf{x}\right)\,d\mathbf{x}=1$.
The Bayes optimal decision at point $\mathbf{x}$ is given by
\[
h_{\mathrm{Bayes}}\left(\boldsymbol{\mathsf{o}}\right)=\textrm{sign}\left(\boldsymbol{\mathsf{f}}_{\mathsf{X}|\mathsf{Y}=+1}\left(\mathbf{x}_{\boldsymbol{\mathsf{o}}}\right)-\boldsymbol{\mathsf{f}}_{\mathsf{X}|\mathsf{Y}=-1}\left(\mathbf{x}_{\boldsymbol{\mathsf{o}}}\right)\right)
\]
 and can be approximated by
\[
\widehat{h}_{\sigma}\left(\boldsymbol{\mathsf{o}}\right)=\textrm{sign}\left(\boldsymbol{\widehat{\mathsf{f}}}_{\mathsf{X}|\mathsf{Y}=+1}\left(\mathbf{x}_{\boldsymbol{\mathsf{o}}}\right)-\boldsymbol{\widehat{\mathsf{f}}}_{\mathsf{X}|\mathsf{Y}=-1}\left(\mathbf{x}_{\boldsymbol{\mathsf{o}}}\right)\right)\,.
\]
This estimator is shown in Figure \ref{fig:1nn} for the RBF-kernel
(\ref{eq:rbf}) and three different values of $\sigma$. The convergence
$\lim_{\sigma\rightarrow0}\widehat{h}_{\sigma}\left(\boldsymbol{\mathsf{o}}\right)=\mathrm{NN}_{1}\left(\boldsymbol{\mathsf{o}}\right)$
leads to the convergence 
\[
\lim_{\sigma\rightarrow0}R\left(\widehat{h}_{\sigma}\right)=R\left(\mathrm{NN}_{1}\right)
\]
 on which our analysis is based. Note that the performance of $\widehat{h}_{\sigma}$
for small sample size may be bad. Also for increasing sample size
$m\rightarrow\infty$ a a decreasing kernel bandwidth $\sigma\rightarrow0$
is required for consistency. 

Let us consider $k_{0}\left(\boldsymbol{\mathsf{o}},\boldsymbol{\mathsf{o}}_{i}\right)\equiv\lim_{\sigma\rightarrow0}k_{\sigma}\left(\boldsymbol{\mathsf{o}},\boldsymbol{\mathsf{o}}_{i}\right)$
and classifiers of the form
\begin{equation}
g_{\boldsymbol{\alpha}}\left(\boldsymbol{\mathsf{o}}\right)=\textrm{sign}\left(\sum_{i=1}^{m}\alpha_{i}k_{0}\left(\boldsymbol{\mathsf{o}},\boldsymbol{\mathsf{o}}_{i}\right)\right)\,.\label{eq:kernel_class}
\end{equation}
The $1$-NN classifier is then given by $\mathrm{NN}_{1}\left(Z,\cdot\right)=g_{\boldsymbol{\alpha}=\mathbf{y}}\equiv g_{\mathbf{y}}$,
and can be expressed if we restrict the coefficients $\alpha_{i}$
to take values only from $A=\mathcal{Y}=\left\{ -1,+1\right\} $.
Hence, the resulting hypothesis space is given by
\begin{equation}
\mathcal{G}=\left\{ g_{\boldsymbol{\alpha}}:\forall i\in\left\{ 1,\ldots,m\right\} :\alpha_{i}\in A\right\} \,.\label{eq:hypo_bin}
\end{equation}
It turns out that the $1$-NN classifier $g_{\mathbf{y}}$ is a minimiser
of the empirical risk (\ref{eq:r_emp}), i.e.~$g_{\mathbf{y}}=\mathrm{argmin}_{g\in\mathcal{G}}\,R_{\mathrm{emp}}\left(g,Z\right)$.
This is easily seen by considering that $\forall\boldsymbol{\mathsf{o}}\in O$
the $1$-neighbourhood $N_{1}\left(\boldsymbol{\mathsf{o}}\right)=\left\{ \boldsymbol{\mathsf{o}}\right\} $
and thus $\forall\left(\boldsymbol{\mathsf{o}},y\right)\in Z:g_{\mathbf{y}}\left(\boldsymbol{\mathsf{o}}\right)=y$
resulting in $R_{\mathrm{emp}}\left(g_{\mathbf{y}},Z\right)=0$. Also,
$g_{\mathbf{y}}\left(\boldsymbol{\mathsf{o}}\right)$ is the only
minimiser of $R_{\mathrm{emp}}\left(g,Z\right)$ because for each
$\alpha_{i}$ flipped, exactly one training error is incurred resulting
in an increase over $R_{\mathrm{emp}}\left(g_{\mathbf{y}},Z\right)$
by $1/m$. Thus the application of $1$-NN conforms to the principle
of \emph{empirical} \emph{risk} \emph{minimisation} \cite{Vapnik:98}
at vanishing training error. 

Based on this view let us turn to an intuitive explanation of why
the $1$-NN classifier is able to generalise. As shown in Figure \ref{fig:1nn}
not all the objects $\boldsymbol{\mathsf{o}}\in O$ from the training
sample $O$ contribute to the decision function. In terms of the hypothesis
space defined in (\ref{eq:kernel_class}) and (\ref{eq:hypo_bin})
above this means that the respective summands could be set to nought
without changing the decision at any object $\boldsymbol{\mathsf{o}}\in\mathcal{O}$
. Let us define the set $\overline{Z}_{K}$ of subsets $Z^{\prime}\subset Z$
of training data \emph{redundant} for the $K$-NN classifier by
\[
\overline{Z}_{K}\stackrel{\mathrm{def}}{=}\left\{ Z^{\prime}:\forall\boldsymbol{\mathsf{o}}\in\mathcal{O}\;\mathrm{NN}_{K}\left(Z,\boldsymbol{\mathsf{o}}\right)=\mathrm{NN}_{K}\left(Z\setminus Z^{\prime},\boldsymbol{\mathsf{o}}\right)\right\} ,
\]
 and let $Z_{K}\in\overline{Z}_{K}$ be defined as the element of
$\overline{Z}_{K}$ of maximum cardinality, $Z_{K}=\mathrm{argmax}_{Z^{\prime}\in\overline{Z}_{K}}\left(\left|Z^{\prime}\right|\right).$
Even if all the training examples in $Z_{K}$ were left out the prediction
at no object $\boldsymbol{\mathsf{o}}\in\mathcal{O}$ would change.
Please note the interesting resemblance of $Z\setminus Z_{K}$ to
the set of support vectors in SVM learning \cite{Vapnik:98}. In order
to be able to express this sparseness of solutions in the expansion
coefficients $\alpha_{i}$ let us augment the hypothesis space $\mathcal{G}$
by allowing the coefficients $\alpha_{i}$ to take on nought as an
additional value, $\forall i\in\left\{ 1,\ldots\,m\right\} \;\alpha_{i}\in A\cup\left\{ 0\right\} \equiv\widetilde{A}$.
This will allow us to express prior belief in the sparseness of a
solution by putting additional prior weight on solutions with few
non-vanishing coefficients. The augmented hypothesis space $\widetilde{\mathcal{G}}$
is given by
\begin{equation}
\widetilde{\mathcal{G}}=\left\{ g_{\boldsymbol{\widetilde{\alpha}}}:\alpha_{i}\in\widetilde{A}\right\} .\label{eq:hypo_tern}
\end{equation}
 Then we can define the set $G_{\mathbf{y}}\subseteq\widetilde{\mathcal{G}}$
of hypotheses that are equivalent to $g_{\mathbf{y}}$ w.r.t. the
classification on $\mathcal{O}$ 
\begin{equation}
G_{\mathbf{y}}=\left\{ g_{\boldsymbol{\alpha}}\in\widetilde{\mathcal{G}}:\forall\boldsymbol{\mathsf{o}}\in\mathcal{O}\;g_{\boldsymbol{\alpha}}\left(\boldsymbol{\mathsf{o}}\right)=g_{\mathbf{y}}\left(\boldsymbol{\mathsf{o}}\right)\right\} .\label{eq:equivalence}
\end{equation}
 The cardinality $\left|G_{\mathbf{y}}\right|$ of this set will later
serve as the crucial quantity for bounding the generalisation error.
Since the set $G_{\mathbf{y}}$ is not easily accessible we can define
a subset $G_{Z_{1}}\subseteq G_{\mathbf{y}}$ by
\[
G_{Z_{1}}=\left\{ g_{\boldsymbol{\alpha}}\in\widetilde{\mathcal{G}}:\forall\left(\boldsymbol{\mathsf{o}}_{i},y_{i}\right)\in Z\setminus Z_{1}\quad\alpha_{i}=y_{i}\right\} .
\]
 The cardinality $\left|G_{Z_{1}}\right|$ of this set is then given
by $\left|G_{Z_{1}}\right|=2^{r}$ and is thus trivially related to
the number $r$ of redundant points. It will later serve as a convenient
lower bound, $\left|G_{Z_{1}}\right|\leq\left|G_{\mathbf{y}}\right|$.
The redundancy $r$ can also be viewed as a kind of luckiness in the
sense of \cite{Taylor:96}. 

\section{A PAC-Bayesian bound for 1-NN\label{sec:bound}}

We would like to define a prior over $\widetilde{\mathcal{G}}$ and
apply the PAC-Bayesian Theorem \ref{thm:dma1}. However, the prior
over the hypothesis space $\mathcal{H}$ as referred to in Theorem
\ref{thm:dma1} requires us to define an hypothesis space $\mathcal{H}$
\emph{before} learning. In contrast, the hypothesis space $\widetilde{\mathcal{G}}$
defined by equations (\ref{eq:kernel_class}) and (\ref{eq:hypo_tern})
appears to be data-dependent and thus not known before the data are
considered. Let us consider an alternative hypothesis space given
by all the linear functions 
\[
\mathcal{H}=\left\{ h_{\mathbf{w}}:h_{\mathbf{w}}=\mathrm{sign}\left(\left\langle \mathbf{w},\boldsymbol{\phi}\left(\boldsymbol{\mathsf{o}}\right)\right\rangle _{\mathcal{K}}\right),\mathbf{w}\in\mathcal{K},\left\Vert \mathbf{w}\right\Vert _{\mathcal{K}}=1\right\} .
\]
$\mathcal{K}$ is the kernel space associated with the kernel 
\[
k_{\sigma}\left(\boldsymbol{\mathsf{o}},\boldsymbol{\mathsf{o}}^{\prime}\right)=\left\langle \boldsymbol{\phi}\left(\boldsymbol{\mathsf{o}}\right),\boldsymbol{\phi}\left(\boldsymbol{\mathsf{o}}^{\prime}\right)\right\rangle _{\mathcal{K}},
\]
and $\boldsymbol{\phi}:\mathcal{O}\mapsto\mathcal{K}$. The unit length
constraint $\left\Vert \mathbf{w}\right\Vert _{\mathcal{K}}=1$ is
required in order to be able to define a proper (normaliseable) prior
measure over $\mathcal{H}_{\mathbf{w}}$ such that $\boldsymbol{\mathsf{P}}_{\mathsf{H}_{\mathbf{w}}}\left(\mathcal{H}_{\mathbf{w}}\right)=1$.
Since we can expand the weight vector $\mathbf{w}$ in terms of the
objects $\boldsymbol{\mathsf{o}}_{i}\in Z$ by 
\begin{equation}
\mathbf{w}=\sum_{i=1}^{m}\alpha_{i}\boldsymbol{\phi}\left(\boldsymbol{\mathsf{o}}_{i}\right)\label{eq:expansion}
\end{equation}
the hypotheses as given in equation (\ref{eq:kernel_class}) can be
written as
\begin{eqnarray*}
g_{\boldsymbol{\alpha}}\left(\boldsymbol{\mathsf{o}}\right) & = & \mathrm{sign}\left(\sum_{i=1}^{m}\alpha_{i}k_{\sigma}\left(\boldsymbol{\mathsf{o}},\boldsymbol{\mathsf{o}}_{i}\right)\right)\\
 & = & \mathrm{sign}\left(\sum_{i=1}^{m}\alpha_{i}\left\langle \boldsymbol{\phi}\left(\boldsymbol{\mathsf{o}}_{i}\right),\boldsymbol{\phi}\left(\boldsymbol{\mathsf{o}}\right)\right\rangle _{\mathcal{K}}\right)\\
 & = & \mathrm{sign}\left(\left\langle \mathbf{w},\boldsymbol{\phi}\left(\boldsymbol{\mathsf{o}}\right)\right\rangle _{\mathcal{K}}\right)\\
 & = & h_{\mathbf{w}}\left(\boldsymbol{\mathsf{o}}\right)
\end{eqnarray*}
Thus for every hypothesis $g_{\boldsymbol{\alpha}}\in\widetilde{\mathcal{G}}$
there exists a corresponding hypothesis $h_{\mathbf{w}}\in\mathcal{H}$
\emph{before} the training data $Z$ are considered. Since Theorem
\ref{thm:dma1} holds for any two probability measures $\boldsymbol{\mathsf{P}}_{\mathsf{H}}$
and $\boldsymbol{\mathsf{P}}_{\mathsf{OY}}$ it is sufficient to show
that given any prior measure $\boldsymbol{\mathsf{P}}_{\widetilde{\mathsf{G}}}$
over $\widetilde{\mathcal{G}}$ there always exists a corresponding
prior measure $\boldsymbol{\mathsf{P}}_{\mathsf{H}}$ over $\mathcal{H}$. 

Let us define the $\left(\dim\left(\mathcal{K}\right)\times m\right)$-matrix
\[
\boldsymbol{\Phi}\left(O\right)=\left(\boldsymbol{\phi}\left(\boldsymbol{\mathsf{o}}_{1}\right),\ldots,\boldsymbol{\phi}\left(\boldsymbol{\mathsf{o}}_{m}\right)\right)
\]
 of training objects $\boldsymbol{\mathsf{o}}$ mapped to kernel space
$\mathcal{K}$. Then the linear transformation from the parameter
space $\widetilde{A}^{m}$ to kernel space \emph{$\mathcal{K}$} can
be written as $\mathbf{w}=\boldsymbol{\Phi}\left(O\right)\boldsymbol{\alpha}$
and we have for any measurable subset $H\subseteq\mathcal{H}$ a corresponding
set $\widetilde{G}\subseteq\widetilde{\mathcal{G}}$ given by 
\[
\widetilde{G}\left(H,O\right)=\left\{ g_{\boldsymbol{\alpha}}:\exists\mathbf{w}\in H\quad\frac{\boldsymbol{\Phi}\left(O\right)\boldsymbol{\alpha}}{\left\Vert \boldsymbol{\Phi}\left(O\right)\boldsymbol{\alpha}\right\Vert }=\mathbf{w}\right\} 
\]
The resulting prior measure $\boldsymbol{\mathsf{P}}_{\mathsf{H}}$
is given by 
\[
\boldsymbol{\mathsf{P}}_{\mathsf{H}}\left(H\right)=\boldsymbol{\mathsf{E}}_{\mathsf{O}^{m}}\left[\boldsymbol{\mathsf{P}}_{\widetilde{\mathsf{G}}}\left(\widetilde{G}\left(H,\mathsf{O}\right)\right)\right],
\]
indicating that knowledge of the measure $\boldsymbol{\mathsf{P}}_{\mathsf{O}}$
over objects is necessary in order to determine $\boldsymbol{\mathsf{P}}_{\mathsf{H}}$.
This does not constitute a problem, however, because explicit knowledge
of $\boldsymbol{\mathsf{P}}_{\mathsf{H}}$ is neither required for
the application of the algorithm nor for the calculation of the PAC-Bayesian
bound values.

First, let us illustrate the application of the PAC-Bayesian bound
(\ref{thm:dma1}) by constructing a very simple prior $\boldsymbol{\mathsf{P}}_{\widetilde{\mathsf{G}}}\left(\boldsymbol{\alpha}\right)$
over $\widetilde{\mathcal{G}}$. Due to the iid property of the training
sample $Z$, we have no knowledge about any specific $\alpha_{i}$
and thus choose a factorising prior 
\begin{equation}
\boldsymbol{\mathsf{P}}_{\widetilde{\mathsf{G}}}\left(g_{\boldsymbol{\alpha}}\right)=\prod_{i=1}^{m}\boldsymbol{\mathsf{P}}_{\widetilde{\mathsf{A}}}\left(\alpha_{i}\right),\label{eq:factorise}
\end{equation}
that reflects the interchangeability of the training examples in $Z$.
Assuming no further knowledge about the plausibility of hypotheses
let us choose the prior to be uniform,
\[
\boldsymbol{\mathsf{P}}_{\widetilde{\mathsf{A}}=-1}\left(\alpha_{i}\right)=\boldsymbol{\mathsf{P}}_{\widetilde{\mathsf{A}}=1}\left(\alpha_{i}\right)=\boldsymbol{\mathsf{P}}_{\widetilde{\mathsf{A}}=0}\left(\alpha_{i}\right)=\frac{1}{3},
\]
which obviously leads to a uniform measure $\boldsymbol{\mathsf{P}}_{\widetilde{\mathsf{G}}}\left(\boldsymbol{\alpha}\right)$,
as well. This choice will later be refined in the light of general
knowledge about the sparseness of typical $1$-NN classifiers. Then
the measure of hypotheses $g_{\boldsymbol{\alpha}}\in G_{\mathbf{y}}$
equivalent to $g_{\mathbf{y}}$ on $\mathcal{O}$ is given by \hfill{}
\begin{equation}
\boldsymbol{\mathsf{P}}_{\widetilde{\mathsf{G}}}\left(G_{\mathbf{y}}\right)=\frac{\left|G_{\mathbf{y}}\right|}{\left|\widetilde{\mathcal{G}}\right|}\geq\frac{\left|G_{Z_{1}}\right|}{\left|\widetilde{\mathcal{G}}\right|}=\frac{2^{r}}{3^{m}-1},\label{eq:simple_prior}
\end{equation}
because among the total of $3^{m}-1$ hypotheses in $\widetilde{\mathcal{G}}$
we have $2^{r}$ hypotheses that agree with $g_{\mathbf{y}}$ on \emph{$\mathcal{O}$}.
Then we can give the following bound on the generalisation error of
$1$-NN.
\begin{thm}
\label{thm:1nn_bound}For any probability distribution $\boldsymbol{\mathsf{P}}_{\mathsf{Z}}$
on labelled objects we have, for any $\delta>0,$ that with probability
at least $1-\delta$ over the selection of a sample of $m$ examples,
the following holds for the $1$-NN classifier $g_{\mathbf{y}}$ with
$r$ redundant examples :
\[
R\left(g_{\mathbf{y}}\right)\leq\frac{m\log3-r\log2+\log\frac{1}{\delta}}{m}.
\]
\end{thm}
\begin{figure}
\begin{centering}
\includegraphics[angle=270,width=0.9\columnwidth]{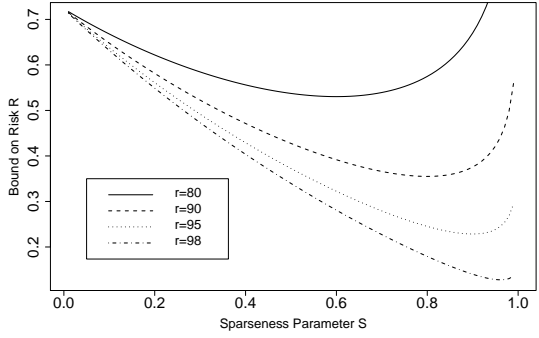}
\par\end{centering}
\caption{\label{fig:bound_R_S}Values of the bound given in Theorem \ref{thm:1nn_bound2}
as a function of the expected sparsity $S$ for four different values
$r$ of the observed number of redundant objects. The training set
size is $m=100$ and the confidence is $95\%$ corresponding to $\delta=0.05$.
Large values of $r$ lead to lower values of the bound, but the bound
attains its minimum only if the expected sparsity $S$ matches the
the number of redundant objects $r$. Note that the optimum $S_{\mathrm{opt}}$
for a given redundancy $r$ is $S_{\mathrm{opt}}<\frac{r}{m}$, the
value one may have expected.}
\end{figure}
Let us refine this bound by constructing a more informative prior
$\boldsymbol{\mathsf{P}}_{\widetilde{\mathsf{G}}}$. Maintaining the
factorising property (\ref{eq:factorise}) and introducing an expected
level $S$ of sparsity we choose
\[
\boldsymbol{\mathsf{P}}_{\widetilde{\mathsf{A}}=0}\left(\alpha_{i}\right)=S\;\mathrm{and}\;\boldsymbol{\mathsf{P}}_{\widetilde{\mathsf{A}}=-1}\left(\alpha_{i}\right)=\boldsymbol{\mathsf{P}}_{\widetilde{\mathsf{A}}=1}\left(\alpha_{i}\right)=\frac{1-S}{2}
\]
The resulting prior measure $\boldsymbol{\mathsf{P}}_{\widetilde{\mathsf{G}}}$
is then only a function of the sparsity $s\left(g_{\boldsymbol{\alpha}}\right)$
of an hypothesis $g_{\boldsymbol{\alpha}}$ given by $s\left(g_{\boldsymbol{\alpha}}\right)=\left|\left\{ i\in\left\{ 1,\ldots\,m\right\} :\alpha_{i}=0\right\} \right|$.
We are interested in the prior measure $\boldsymbol{\mathsf{P}}_{\widetilde{\mathsf{G}}}\left(G_{\mathbf{y}}\right)$
of all those hypotheses $g_{\boldsymbol{\alpha}}\in\widetilde{\mathcal{G}}$
that behave equivalently to $g_{\mathbf{y}}$ on \emph{$\mathcal{O}$.}
This quantity is lower bounded by $\boldsymbol{\mathsf{P}}_{\widetilde{\mathsf{G}}}\left(G_{Z_{1}}\right)$.
\begin{eqnarray}
\boldsymbol{\mathsf{P}}_{\widetilde{\mathsf{G}}}\left(G_{Z_{r}}\right) & = & \sum_{s=0}^{r}\left(\begin{array}{c}
r\\
s
\end{array}\right)\frac{S^{s}\left(\frac{1-S}{2}\right)^{m-s}}{1-S^{m}}\nonumber \\
 & = & \frac{\left(\frac{1-S}{2}\right)^{m-r}}{1-S^{m}}\sum_{s=0}^{r}\left(\begin{array}{c}
r\\
s
\end{array}\right)S^{s}\left(\frac{1-S}{2}\right)^{r-s}\nonumber \\
 & = & \frac{\left(\frac{1-S}{2}\right)^{m-r}\left(\frac{1}{2}\left(1+S\right)\right)^{r}}{1-S^{m}}\nonumber \\
 & = & \frac{\left(1-S\right)^{m-r}\left(1+S\right)^{r}}{2^{m}\left(1-S^{m}\right)}.\label{eq:refined_prior}
\end{eqnarray}

\begin{figure}
\begin{centering}
\includegraphics[angle=270,width=0.9\columnwidth]{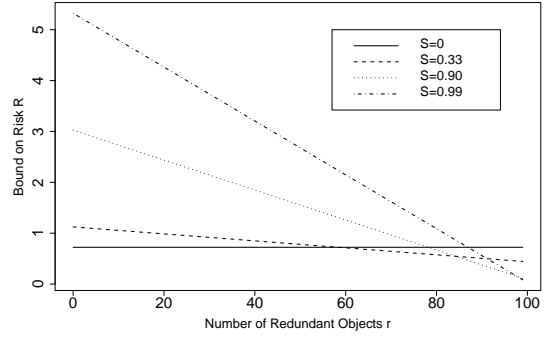}
\par\end{centering}
\caption{\label{fig:bound_R_r}Values of the bound given in Theorem \ref{thm:1nn_bound2}
as a function of the number $r$ of redundant objects for four different
values of the expected sparsity $S$. The training set size is $m=100$
and the confidence is $95\%$ corresponding to $\delta=0.05$. Large
values of $S$ lead to lower values of the bound, but only for sufficiently
large values of $r$. If the value of $S$ is chosen too optimistically,
the resulting value of the bound suffers.}
\end{figure}

Note, that this reduces to the previous result (\ref{eq:simple_prior})
for $S=1/3.$ Using the result (\ref{eq:refined_prior}) we can give
a more refined PAC-Bayesian bound on the generalisation error of $1$-NN. 
\begin{thm}
\label{thm:1nn_bound2}For any distribution $\boldsymbol{\mathsf{P}}_{\mathsf{Z}}$
over labelled objects and any sparsity value $S\in\left[0,1\right[$
chosen a-priori, we have, for any $\delta>0,$ that with probability
at least $1-\delta$ over the selection of a sample of $m$ examples,
the following holds for the $1$-NN classifier $g_{\mathbf{y}}$ with
$r$ redundant examples :
\[
R\left(g_{\mathbf{y}}\right)\leq\frac{m\log\frac{2\left(1-S^{m}\right)}{\left(1-S\right)}+r\log\frac{\left(1-S\right)}{\left(1+S\right)}+\log\frac{1}{\delta}}{m}.
\]
\end{thm}
In order to get a feel for the bound, consider first Figure \ref{fig:bound_R_S}.
The convex shapes of the curves clearly indicate that a wrong choice
of $S$ hurts in both cases: For over- and underestimated redundancy.
Figure \ref{fig:bound_R_r} illustrates the behaviour of the bound
as a function of redundancy $r$. The case $S=0$ effectively corresponds
to the unaugmented hypothesis space $\mathcal{G}$ with a flat prior.
Due to the increase $\left|\mathcal{G}\right|=2^{m}$ of $\mathcal{G}$
with $m$ the resulting cardinality bound can never give values below
$\sqrt{2}\approx0.69$. The case $S=0.33$ corresponds to the bound
of Theorem \ref{thm:1nn_bound} and is superior mostly in ``trivial''
regimes with $R>0.5$. Only for ``courageous'' choices of $S=0.9$
and $S=0.99$ does the bound reach non-trivial regimes. It should
be noted that standard VC-bounds often require training set sizes
of $m>100000$ for even the luckiest cases to give non-trivial bounds
($R<0.5$).

As a matter of fact, it is feasible to incorporate even more knowledge
than the level of sparsity $S$ into the bound. In addition, knowledge
about the a-priori class probabilities $\boldsymbol{\mathsf{P}}_{\mathsf{Y}}\left(\mathsf{Y}=\pm1\right)$
and knowledge about the levels of sparsity $S^{\pm}$ in each of the
classes could be incorporated in the bound. 

\section{The general case of $K$-NN\label{sec:knn}}

In practice, people often use the $K$-NN classifier, $K>1$, rather
than the $1$-NN classifier to avoid over-fitting the data. In order
to arrive at a similar result as that obtained in Section \ref{sec:bound}
let us find a formulation for $K$-NN equivalent to that given in
(\ref{eq:1nn}) for $1$-NN. We avoid the problem of voting ties by
considering only odd values of $K$. Since the $K$ nearest neighbours
need to be selected, we use a product of kernels,

\begin{eqnarray}
\mathrm{NN}_{K}\left(Z,\boldsymbol{\mathsf{o}}\right) & = & \mathrm{sign}\left(\sum_{\stackrel{\left(\boldsymbol{\mathsf{o}}^{\prime},y^{\prime}\right)\in Z:}{\boldsymbol{\mathsf{o}}^{\prime}\in N_{K}(\boldsymbol{\mathsf{o}})}}y^{\prime}\right)\nonumber \\
 & = & \mathrm{sign}\left(\sum_{\stackrel{Z^{\prime}\subseteq Z:}{\left|Z^{\prime}\right|=K}}\prod_{Z^{\prime}}k_{0}\left(\boldsymbol{\mathsf{o}},\boldsymbol{\mathsf{o}}^{\prime}\right)\sum_{Z^{\prime}}^{K}y^{\prime}\right)\nonumber \\
 & = & \mathrm{sign}\left(\sum_{\mathbf{i}\in I}\prod_{j=1}^{K}k_{0}\left(\boldsymbol{\mathsf{o}},\boldsymbol{\mathsf{o}}_{i_{j}}\right)\sum_{l=1}^{K}y_{i_{l}}\right).\label{eq:knn}
\end{eqnarray}
 The sum is over the set $I$ of index vectors $\mathbf{i}\in I$
defined as 
\[
I\equiv\left\{ \mathbf{i}\in\left\{ 1,\ldots,m\right\} ^{K}:\forall j\in\left\{ 1,\ldots,m-1\right\} \:i_{j+1}>i_{j}\right\} ,
\]
 and we use components $i_{j}$ of $\mathbf{i}=\left(i_{1},\ldots,i_{K}\right)^{\prime}$
for indexing. Again the above classifier can be considered a linear
classifier in a kernel space if we define an augmented product kernel
$\widetilde{k}:\mathcal{O}^{K}\times\mathcal{O}^{K}\mapsto\mathbb{R}^{+}$by
\[
\widetilde{k}\left(\boldsymbol{\mathsf{o}}_{1},\ldots\,\boldsymbol{\mathsf{o}}_{K},\boldsymbol{\mathsf{o}}_{K+1},\ldots,\boldsymbol{\mathsf{o}}_{2K}\right)\equiv\prod_{j=1}^{K}k\left(\boldsymbol{\mathsf{o}}_{j},\boldsymbol{\mathsf{o}}_{j+K}\right).
\]
The product kernel $\widetilde{k}$ retains its Mercer property due
to the closure of kernels under the tensor product \cite{Haussler:99}.
Defining coefficients $\beta_{\mathbf{i}}\equiv\sum_{l=1}^{K}\alpha_{i_{l}}$
with $\forall i\in\left\{ 1,\ldots,m\right\} \:\alpha_{i}=y_{i}$
we express the $K$-NN classifier as the limiting case of a linear
classifier
\[
\mathrm{NN}_{K}\left(Z,\boldsymbol{\mathsf{o}}\right)=\mathrm{sign}\left(\sum_{\mathbf{i}\in I}\beta_{\mathbf{i}}\widetilde{k}_{0}(\underbrace{\boldsymbol{\mathsf{o}},\ldots,\boldsymbol{\mathsf{o}}}_{K\,\mathrm{times}},\boldsymbol{\mathsf{o}}_{i_{1}},\ldots,\boldsymbol{\mathsf{o}}_{i_{j}})\right).
\]
Since the coefficients $\beta_{\mathbf{i}}=\beta_{\mathbf{i}}\left(\boldsymbol{\alpha}\right)$
are fully determined by the values of the $\alpha_{i}$ it is sufficient
to consider these. As discussed in Section \ref{sec:1nn} the $1$-NN
classifier can be considered as the \emph{empirical} \emph{risk} \emph{minimiser}
with vanishing training error. The situation is different for $K$-NN.
Consider, e.g.~the situation of a training sample of three different
objects two of which belong to one class and one of which belongs
to the other class. For $K=3$ under any metric the $3$-NN classifier
will incur a loss of $1/3$ because the single object belonging to
the minority class will be classified as belonging to the majority
class. 

Again we can use the redundancy of features to benefit from sparse
solutions in the coefficients $\alpha_{i}$. As in the case of $1$-NN
the two types of results as in Theorems \ref{thm:1nn_bound} and \ref{thm:1nn_bound2}
are possible, this time base on Theorem \ref{thm:dma2}. We will give
here only the version corresponding to Theorem \ref{thm:1nn_bound2}
because Theorem \ref{thm:1nn_bound} follows as a special case thereof.
\begin{thm}
\label{thm:knn_bound}For any probability distribution $\boldsymbol{\mathsf{P}}_{\mathsf{Z}}$
on labelled objects and any sparsity value $S\in\left[0,1\right[$
chosen a-priori, we have, for any $\delta>0,$ that with probability
at least $1-\delta$ over the selection of a sample of $m$ examples,
the following holds for the $K$-NN classifier $g_{\mathbf{y}}^{K}$
with $r_{K}=\log_{2}\left|G_{Z_{K}}\right|$ redundant examples: The
difference 
\[
\Delta R=R\left(g_{\mathbf{y}}^{K}\right)-R_{\mathrm{emp}}\left(g_{\mathbf{y}}^{K};Z\right)
\]
 between actual and empirical risk is bounded from above by
\[
\Delta R\leq\sqrt{\frac{m\log\frac{2\left(1-S^{m}\right)}{\left(1-S\right)}+r_{K}\log\frac{\left(1-S\right)}{\left(1+S\right)}+\log\frac{1}{\delta}}{2m}}.
\]
\end{thm}
While this bound behaves similarly to the one given in Theorem \ref{thm:1nn_bound2}
in terms of $r_{K}$ and $S$, it is more interesting to ask about
the dependency of $\left|G_{\mathbf{y}}\right|$ (or its lower bound
$2^{r_{K}}$) on the number $K$ of neighbours considered. Empirical
results indicate that the risk $R(g_{\mathbf{y}}^{K})$ is a bowl-shaped
function of $K$, indicating the existence of an optimum number $K>1$.
A corresponding theoretical result together with Theorem \label{thm:knn_bound}
would then yield a sound explanation of why $K>1$ may be preferred,
and may even serve as a guide for model selection.

\section{Conclusions and Future Work\label{sec:conclusion}}

We provided small sample size bounds on the generalisation error of
the $K$-nearest-neighbour classifier in the PAC-Bayesian framework
by viewing $K$-NN as a linear classifier in a collapsed kernel space.
Referring back to the goal set in the Introduction these bounds may
serve to make $K$-NN a self-bounding algorithm in the sense of \cite{Freund:98}.
It is left for future research to provide means for determining \emph{in}
\emph{practice} at least an estimate of the number $r_{K}$ of redundant
points. 

Interestingly, our analysis involves the notion of redundant examples
and --- as a consequence --- of essential examples that bear a close
resemblance with \emph{support} \emph{vectors} \cite{Vapnik:98}.
Also, considering Figure \ref{fig:1nn} it is obvious that $1$-NN
performs a \emph{local} \emph{margin} \emph{maximisation} as opposed
to a global margin maximisation in the SVM.

Pursuing the similarity to SVMs further, note that the $K$-NN classifier
not only returns a classification for a given object $\boldsymbol{\mathsf{o}}\in\mathcal{O}$,
but also provides a discrete margin 
\[
\gamma\left(\boldsymbol{\mathsf{o}}\right)=y\sum_{i:\boldsymbol{\mathsf{o}}_{i}\in N_{K}(\boldsymbol{\mathsf{o}})}y_{i}
\]
 taking values $\gamma\in\left\{ -K,-K+2,\ldots,K-2,K\right\} $.
Hence, we can define the margin $\gamma_{Z}$ on the training sample
$Z$ by $\gamma_{Z}=\min_{\left(\boldsymbol{\mathsf{o}},y\right)\in Z}\gamma\left(\boldsymbol{\mathsf{o}}\right).$
Since the now famous Support Vector Machine \cite{Vapnik:98} is based
on maximising the margin and also generalisation bounds for linear
classifiers \cite{Herbrich:99b,Vapnik:98} are based on this notion
it is tempting to speculate that also for $K$-NN the margin $\gamma_{Z}$
on the training sample should play a role in the generalisation bound.
Intuitively, the relation between $\left|G_{\mathbf{y}}\right|$ and
the margin $\gamma_{Z}$ is clear: The more unanimous the outcome
of the voting on $O$ the more hypotheses $g_{\boldsymbol{\alpha}}\in\widetilde{\mathcal{G}}$
would give the same classification of the training data and therefore
more likely agree on $\mathcal{O}$. However, at this point it is
not clear how exactly the margin $\gamma_{Z}$ is related to $\left|G_{\mathbf{y}}\right|$,
the quantity determining generalisation.

Another interesting aspect of the margin $\gamma\left(\boldsymbol{\mathsf{o}}\right)$
is its use as a confidence measure for the prediction of labels on
test objects. For linear classifiers this method has been theoretically
justified by \cite{Taylor:96a}. Indeed, for $K$-NN such a strategy
has been put forward in the form of the $\left(K,L\right)$-nearest-neighbour
rule \cite{Hellman:70} that given a parameter $L>K/2$ refuses to
make predictions at $\boldsymbol{\mathsf{o}}$ unless $\gamma\left(\boldsymbol{\mathsf{o}}\right)\geq L$.
Depending on $L$ this principle leads to a rejection rate $\rho\left(L\right)$
on a given test sample. Based on $L$ and $\rho\left(L\right)$ it
should be possible to bound the risk on the non-rejected points. 

\bibliographystyle{chicago}
\bibliography{colt,stat}

\end{document}